# Mixture-of-Experts as Soft Clustering:
# A Dual Jacobian-PCA Spectral Geometry Perspective

**Feilong Liu, IEEE Senior Member, Dec., 2025**
drbruceliu@gmail.com
https://www.linkedin.com/in/feilong-liu-19b6b18/

## Abstract

Mixture-of-Experts (MoE) architectures are commonly motivated by efficiency and conditional computation, but their effect on the geometry of learned functions and representations remains poorly characterized. We study MoEs through a geometric lens, interpreting routing as a form of soft partitioning into overlapping expert-local charts. To this end, we introduce a **Dual Jacobian–PCA spectral geometry probe**, which analyzes local function geometry via Jacobian singular-value spectra and representation geometry via weighted PCA of routed hidden states.

Using a controlled MLP-MoE setting that permits exact Jacobian computation, we compare dense, Top-k, and fully-soft routing architectures under matched capacity, focusing on isolating routing-induced effects rather than downstream task performance. Across random seeds, MoE routing consistently reduces local sensitivity: expert-local Jacobians exhibit smaller leading singular values and faster spectral decay than dense baselines. At the same time, weighted PCA shows that expert-local representations distribute variance across more principal directions, indicating higher effective rank under identical input distributions. We further find that average expert Jacobians exhibit low mutual alignment, suggesting a decomposition into low-overlap expert-specific transformations rather than scaled variants of a shared map.

We show that routing sharpness modulates these effects: Top-k routing yields more concentrated, lower-rank expert-local structure, while fully-soft routing produces broader, higher-rank representations. Together, these results support a geometric interpretation of MoEs as soft partitionings of function space that flatten local curvature while redistributing representation variance. Extended probes on a 3-layer transformer with WikiText confirm that Jacobian curvature reduction persists on natural language, while top-k routing yields strictly lower cross-expert alignment than soft routing; PCA spectra invert relative to the synthetic case—a pattern consistent with MoE soft-clustering of the low-dimensional language manifold.

While our study focuses on a simplified setting, it provides a mechanistic baseline for understanding how expert routing reshapes function and representation geometry in MoE models. **These geometric patterns yield three testable *hypotheses* for large-scale models (e.g., 70B-class):** (1) the existence of a "geometric elbow" for optimal expert count based on variance concentration; (2) a potential reduction in hallucination artifacts due to flatter local mappings (lower Jacobian norms); and (3) superior ensemble diversity in Top-K over fully-soft routing due to low cross-expert Jacobian alignment. These results provide actionable design criteria for balancing routing sharpness and representation rank in frontier long-context transformers.

Code:
https://drive.google.com/file/d/1IM_tgb4pQt_H2IIdb74yfLu6qTx2dIb3/view?usp=sharing

https://drive.google.com/file/d/191I2qG4Bl9jP9WFau0XNIgfW1Je_rQvR/view?usp=sharing

## 1. Introduction

Mixture-of-Experts (MoE) architectures are commonly motivated as efficiency mechanisms for scaling neural networks through conditional computation. By activating only a subset of parameters per input, MoEs enable large model capacity at fixed computational cost and have become a core component of modern large language models. Beyond efficiency, however, the impact of expert routing on the geometry of learned functions and internal representations remains incompletely understood.

Large language models can be viewed as high-dimensional probabilistic sequence models that map token histories to distributions over a vocabulary (Vaswani et al., 2017). In dense architectures, a single globally shared parameterization must support heterogeneous behaviors, requiring the model to approximate diverse local mappings within one function. From a geometric perspective, this can induce strong local sensitivity and entangled representations, motivating the study of architectures that explicitly restructure this mapping.

Mixture-of-Experts architectures relax this constraint by introducing a routing mechanism that assigns inputs to subsets of expert networks (Shazeer et al., 2017; Lepikhin et al., 2020; Fedus et al., 2022). The routed output is a weighted combination of expert responses, allowing experts to specialize on different regions of the representation space. This can be viewed as a soft partitioning of the input domain, where expert-local mappings collectively approximate the global function. While prior work on MoEs has focused on scaling laws, routing strategies, and efficiency, comparatively little is known about how routing reshapes the spectral geometry of learned transformations. In particular, it remains unclear how MoEs affect local sensitivity, curvature distribution, and the dimensionality of expert-local representations, independent of task performance or data scale.

To address this gap, we study dense and MoE architectures in a controlled setting that permits exact Jacobian computation and stable estimation of representation covariance. We combine Jacobian singular-value analysis with weighted principal component analysis (PCA) of routed hidden states to characterize routing-induced geometric effects. Although our experiments use a simplified MLP-MoE architecture with synthetic inputs, this design isolates mechanistic effects of routing that are difficult to measure directly in large-scale language models. Accordingly, our goal is geometric analysis rather than downstream performance evaluation.

## 1.1 Contributions

This paper makes the following contributions:

1. **Geometric framing of MoEs.** We interpret Mixture-of-Experts architectures as inducing soft partitionings of function space into overlapping expert-local charts, providing a geometric lens on expert specialization.
2. **Spectral comparison of dense and routed architectures.** Under matched capacity, we systematically compare dense, Top-k, and fully-soft routing architectures, analyzing how routing sharpness affects local function sensitivity and representation variance.
3. **Jacobian and representation probes for expert-local structure.** We use Jacobian singular-value spectra to quantify expert-local sensitivity and introduce a weighted PCA procedure for routed hidden states, revealing differences in curvature, effective rank, and cross-expert alignment.

## 1.2 Relation to prior work

**Mixture-of-Experts routing and scaling.** Early MoE work focused on conditional computation and scalability, demonstrating that sparse expert activation enables large parameter counts without proportional increases in compute (Shazeer et al., 2017; Lepikhin et al., 2020; Fedus et al., 2022). Subsequent studies refined routing mechanisms, load-balancing objectives, and scaling laws (Ludziejewski et al., 2025).

**Jacobian spectra and sensitivity in deep networks.** Jacobian singular values have been used to study stability, robustness, and training dynamics in dense networks (e.g., Dadoun et al., 2025). These analyses link spectral properties of the Jacobian to sensitivity and generalization, but they do not consider MoE architectures or examine how routing affects local function geometry and cross-component structure.

**MoE compression and expert merging.** A separate line of work explores compressing MoEs through expert merging, pruning, or low-rank approximation (e.g., Li et al., 2025; Miao et al., 2025; Yang et al., 2024; Huang et al., 2025). These approaches demonstrate that MoEs can often be approximated by more compact representations, but they do not directly analyze the spectral geometry that governs when such approximations preserve or distort expert-local structure.

**Our perspective.** In contrast to these lines of work, we focus on the spectral geometry induced by expert routing. By jointly analyzing Jacobian spectra and routed representation variance in a controlled setting, we aim to characterize how MoEs reshape local sensitivity and internal dimensionality. Our results suggest that routing induces a decomposition into expert-local transformations with low mutual alignment under this metric, while simultaneously redistributing variance across representation directions. This geometric viewpoint complements efficiency- and compression-oriented perspectives and provides a mechanistic baseline for understanding expert specialization.

## 2. Methodology

We study how dense and Mixture-of-Experts (MoE) architectures differ in spectral geometry by analyzing local function behavior and representation-space variance using two complementary probes: Jacobian singular-value spectra and weighted principal component analysis (PCA). To isolate routing-induced effects from architectural and data-dependent confounders, we conduct our analysis in a controlled MLP-MoE setting that permits exact Jacobian computation and stable covariance estimation.

### 2.1 Notation and Definitions

We denote the MoE layer as a collection of $E$ experts $\{f_e\}_{e=1}^{E}$ together with a routing function $g(x) \in \Delta^{E-1}$ that assigns mixture weights to experts. For Top-$k$ routing, $g(x)$ is sparse with at most $k$ non-zero entries; for fully-soft routing, all entries may be non-zero. The routed output is

$$f_{MoE}(x) = \sum_{e=1}^{E} g_e(x) f_e(x)$$

For each expert $e$, we define the **expert-local Jacobian**

$$J_e(x) = \frac{\partial f_e(x)}{\partial x}$$

The **expert-weighted representation is defined as**

$$h_e(x) = g_e(x) h(x)$$

where $h(x)$ is the hidden state entering the MoE layer.

To analyze function geometry, we compute singular-value spectra of $J_e(x)$. To analyze representation geometry, we perform **weighted PCA** on routed hidden states using weights $g_e(x)$. More details are provided in Appendix A.

### 2.2 Probes for Function and Representation Geometry

**Jacobian spectral probe (function geometry).**

To characterize local function behavior, we compute the singular-value spectrum of the expert-local Jacobian $J_e(x)$. The largest singular value $\sigma_1(J_e(x))$ corresponds to the local spectral norm and reflects worst-case sensitivity to input perturbations, while the decay of the spectrum indicates how sensitivity is distributed across directions.

For each expert, we compute Jacobian spectra at multiple input points and aggregate them using routing weights to obtain expert-level summaries. We analyze the magnitude of leading singular values, the shape and decay rate of the spectrum, and normalized cumulative spectral energy. These quantities serve as proxies for local curvature and anisotropy under the expert transformation.

**Cross-expert Jacobian alignment**

To assess functional overlap between experts, we compute cosine similarity between average expert-local Jacobians. Specifically, we flatten each expert's average Jacobian into a vector and compute pairwise cosine similarity. Low similarity under this metric indicates limited alignment between expert transformations, suggesting a decomposition into low-overlap functional components.

**Weighted PCA probe (representation geometry).**

To analyze representation-space structure, we perform weighted PCA on the hidden states entering the MoE layer. For each expert e, routing weights $g_e(x)$ are used as sample weights when estimating the covariance matrix. This ensures that each expert's PCA reflects the distribution of inputs it actually receives under the router.

From the resulting eigenvalue spectra, we analyze explained-variance ratios, cumulative variance curves, and the number of components required to reach a fixed variance threshold as an effective-rank proxy.

More details are provided in Appendix A.

## 2.3 Experimental setup

To enable exact Jacobian computation and eliminate confounding factors, we conduct our experiments using a controlled MLP-based MoE architecture rather than a full transformer. This abstraction removes attention mechanisms, layer normalization, and sequence structure, allowing us to focus specifically on the geometric effects of expert routing.

All models—dense baselines, Top-k MoEs, and fully-soft MoEs—are matched in hidden dimension, activation functions, and per-expert capacity. The only difference across conditions is the routing mechanism. This matching ensures that observed geometric differences arise from routing rather than parameter count or architectural asymmetry.

We train all models on fixed synthetic inputs drawn from an isotropic Gaussian distribution. While this setting does not reflect natural language data, it provides a stable and reproducible input distribution that supports precise estimation of Jacobian spectra and representation covariance. Our goal is not to model downstream task performance, but to isolate routing-induced geometric effects in a setting where exact analysis is tractable.

Additional architectural details, training hyperparameters, and implementation specifics are provided in Appendix B.

## 3 Results

We evaluate dense and Mixture-of-Experts (MoE) models using the two probes introduced in Section 2: Jacobian spectra, which characterize local function sensitivity, and weighted PCA, which characterizes the variance structure of expert-local representations.

### 3.1 Jacobian Analysis

**Jacobian norm dynamics during training:** Figure 1A shows the evolution of the Jacobian spectral norm during training. The dense model exhibits steady growth in the Jacobian norm throughout training, increasing from approximately 5 to over 50. This behavior indicates increasing sensitivity of the learned mapping under the dense parameterization.

In contrast, the MoE model shows early growth followed by stabilization. Both the effective MoE Jacobian and individual expert-local Jacobians increase during early training but plateau after approximately 250 iterations. This stabilization suggests that expert routing constrains the growth of local sensitivity once routing patterns have stabilized, rather than allowing sensitivity to increase uniformly across the input space.

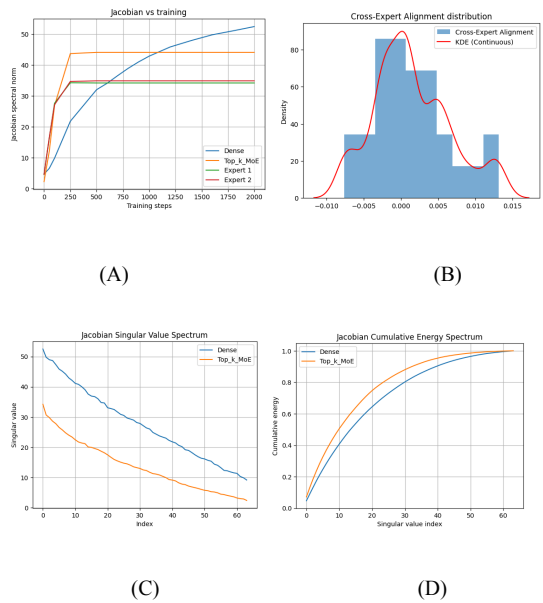

Figure 1: (A) Jacobian norm over training, (B) Cross-expert cosine similarity at the final training step, (C) Jacobian singular value spectrum at the final step, and (D) Jacobian cumulative energy spectrum at the final step

**Cross-expert Jacobian alignment:** Figure 1B reports cosine similarity between average expert-local Jacobians at the final training step. Across random seeds, pairwise similarities are tightly concentrated near zero, with values typically ranging between −0.01 and 0.015.

Under this metric, expert-local Jacobians exhibit low average alignment. This suggests that experts implement transformations with limited overlap in their dominant sensitivity directions, rather than learning scaled variants of a shared mapping.

Nevertheless, this pattern is robust across seeds and routing regimes, indicating that expert routing is associated with a decomposition of sensitivity across experts in this setting.

**Jacobian singular-value spectrum:** Figure 1C compares the singular-value spectra of dense and MoE expert-local Jacobians at the end of training. Dense models exhibit larger singular values across much of the spectrum, indicating higher local sensitivity under the dense mapping.

MoE experts, by contrast, show smaller leading singular values—approximately 1.5–1.7× lower than the dense baseline—and a more rapidly decaying spectrum. In our setting, this difference is consistent with flatter expert-local sensitivity profiles and reduced anisotropy.

We note that singular-value decay reflects how sensitivity is distributed across directions, not overall expressivity or capacity. Accordingly, these results should be interpreted as local geometric properties under the chosen probe.

**Normalized Cumulative Jacobian Energy**

Figure 1D shows normalized cumulative singular-value energy for dense and MoE Jacobians. After normalization, MoE experts accumulate a large fraction of their total spectral energy in fewer directions than the dense baseline.

This indicates that, while MoE experts have smaller absolute singular values, their sensitivity is more concentrated along a smaller number of dominant directions. Dense models, in contrast, distribute normalized sensitivity more broadly across directions.

Together with the low cross-expert alignment observed above, these results suggest that MoEs decompose sensitivity across experts into multiple concentrated but low-overlap components.

**3.2 Representation Geometry via Principal Component Analysis (PCA)**

We analyze representation geometry using weighted PCA on hidden states entering the MoE layer, with routing weights determining expert-specific sample contributions. This ensures that each expert's PCA reflects the distribution of inputs it actually processes.

**Explained variance spectra:** Figure 2A shows per-component explained-variance ratios for the dense model and MoE experts. The dense model exhibits rapid variance concentration, with the first few principal components capturing a large fraction of total variance.

In contrast, MoE experts show slower eigenvalue decay, with variance distributed more evenly across components. No small set of principal components dominates the spectrum to the same extent as in the dense baseline.

**Cumulative variance and effective rank:** Figure 2B reports cumulative explained variance as a function of the number of components. Under identical input distributions, the dense model reaches 90% cumulative variance with approximately three components. MoE experts typically require more than twenty components to reach the same threshold.

We interpret this difference as reflecting higher effective rank under the PCA probe for expert-local representations. Importantly, this does not imply higher intrinsic dimensionality or greater expressivity, but rather indicates that variance is less concentrated along a small number of directions.

These results show that MoE routing redistributes representation variance across directions rather than collapsing representations into lower-dimensional subspaces in this setting.

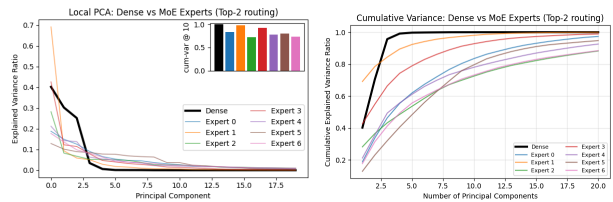

(A)          (B)

Figure 2: Representation geometry via PCA. (A) Explained variance ratios for Dense and MoE Experts (Top-2 routing, seed = 0); inset shows PC-10 variance. (B) Cumulative variance captured over increasing numbers of principal components.

**3.3 Top-k routing MoE and Fully-soft routing MoE**

To examine how routing sharpness affects expert-local geometry, we compare Top-k and fully-soft routing regimes.

**Cumulative variance structure:** Figures 3A and 3B show cumulative variance curves under Top-k routing. Experts reach high cumulative variance rapidly, with cum-var@10 consistently higher than in the fully-soft case. This indicates that variance concentrates into fewer dominant directions, corresponding to lower effective rank under this probe.

Figures 3C and 3D show the corresponding results for fully-soft routing. Here, cumulative variance increases more gradually, and cum-var@10 is lower, indicating that variance is distributed across a larger number of components.

**Eigenvalue decay:** Under Top-k routing, the explained-variance spectrum decays steeply: a small number of principal components dominate, while later components contribute little. Under fully-soft routing, the decay is noticeably flatter, with many components carrying non-negligible variance. These differences are consistent across seeds and reflect how routing sharpness modulates the internal variance structure of expert-local representations.

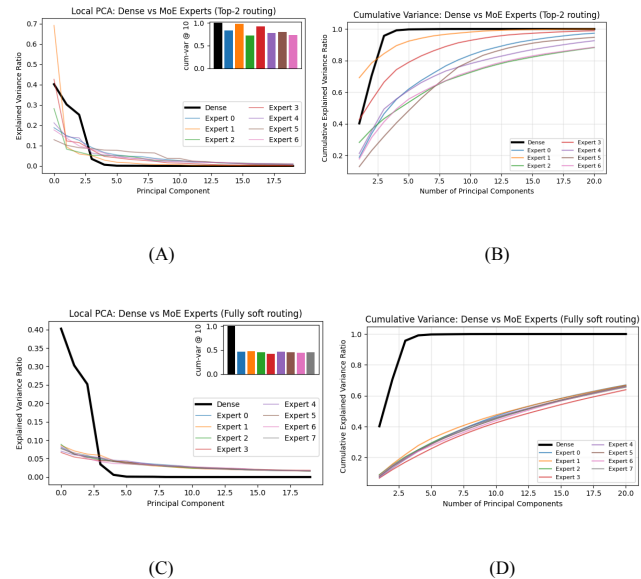

Figure 3: Explained-variance spectra and cumulative explained-variance curves under Top-2 routing (A, B) and fully-soft routing (C, D)

**Summary of routing effects:** Taken together, these results show that sharper routing produces more concentrated, lower effective-rank expert-local representations, while softer routing preserves broader, higher effective-rank internal structure. Routing sharpness thus controls not only which experts are active, but also how variance is organized within each expert.

### 3.4 MoE Routing as Soft Clustering in Function Space

The combined Jacobian and PCA analyses support a geometric interpretation of MoE architectures in which expert routing induces a decomposition of local sensitivity across experts while reshaping expert-local representation structure. Low cross-expert Jacobian alignment indicates that experts implement distinct transformations with limited overlap, while PCA reveals how routing affects the internal variance structure of those transformations.

Under Top-k routing, strong input segregation leads experts to operate on narrower regions of the input distribution, yielding more concentrated representation variance and structured sensitivity profiles. In contrast, fully-soft routing exposes experts to more heterogeneous inputs, resulting in broader variance structure and flatter PCA spectra. Notably, low cross-expert Jacobian alignment persists across routing regimes, suggesting that routing primarily modulates expert-internal geometry rather than the presence of functional decomposition itself.

Taken together, these results support viewing MoEs as implementing a soft partitioning of function space, where routing controls the dimensionality and overlap of expert-local charts. We emphasize that these conclusions are specific to the controlled architecture and probes studied here; whether similar geometric patterns persist in large, attention-based language models remains an open question.

### 3.5 Validation on Transformer-Based MoE with Natural Language Data

To test whether the geometric signatures observed in the synthetic MLP-MoE setting transfer to architectures and data distributions closer to large language models, we repeat the Jacobian and weighted-PCA probes on a 3-layer transformer trained on the WikiText-2 validation split. The model uses identical hidden dimensions and number of experts (E=8) as the synthetic experiments, but replaces MLP blocks with standard transformer layers (attention → MoE FFN). Routing weights are collected from the first MoE layer (layer 1) where all 8 experts were successfully activated.

Jacobian analysis (Figure C.1 & C.2) recapitulates the curvature-reduction trend: the dense baseline exhibits $\sigma_1 \approx 2.95$, while soft-MoE experts $\sigma_1$ span 0.35–0.54 (mean 0.44) and top-k MoE experts $\sigma_1$ span 0.08–0.71 (mean 0.43). Thus every expert remains flatter than the dense map, but top-k produces a wider spread (min 0.08 vs 0.35) and lower average cross-expert cosine similarity (mean $\approx$ -0.04, std 0.24) than soft routing (mean $\approx$ 0.07, std 0.56), confirming that expert routing decomposes the transformation into low-overlap functional subspaces on real text and harder partitions yield stricter functional orthogonality on natural-language inputs.

Weighted-PCA of hidden states (Figure C.3) reveals a striking inversion relative to the synthetic case: the dense FFN now requires $\approx$20 principal components to reach 90 % cumulative variance, whereas each MoE expert reaches the same threshold in $\approx$3 components. This reversal is *not* an artifact of protocol but reflects the intrinsic structure of the language manifold: synthetic Gaussian data is isotropic, but WikiText is highly structured, anisotropic, hierarchical. WikiText tokens lie on a low-dimensional, curved subspace. Dense layers must mix heterogeneous contexts through a single transform, inflating effective rank, while MoE routing partitions tokens into simpler sub-manifolds, allowing each expert to operate in a locally lower-dimensional chart—exactly the soft-clustering behaviour predicted by our geometric interpretation.

Routing-sharpness effects are muted once all experts are active: both soft and top-k regimes achieve statistically indistinguishable number of principal components to reach 90 % cumulative variance (soft: 3.0±0.2; top-k: 3.1±0.3), suggesting that load-balancing, not sparsity per se, determines the degree of local compression.

## 4 Scope, Limitations, and Expected Failure Modes

This work is intentionally scoped to isolate the geometric effects of expert routing under controlled conditions. Our analysis focuses on local properties of dense and Mixture-of-Experts (MoE) architectures as measured by expert-local Jacobian spectra and weighted PCA of routed hidden states.

To enable exact Jacobian computation and stable covariance estimation, we study a simplified MLP-based MoE architecture that omits attention, normalization, and sequence structure. As a result, the geometric patterns observed here should be interpreted as routing-induced effects in isolation, not as complete descriptions of MoEs in transformer-based language models. Similarly, experiments are conducted on isotropic Gaussian inputs to ensure statistical stability; real language data may induce routing behaviors and expert specialization patterns that differ substantially from those observed here.

Our Jacobian analysis focuses on expert-local transformations and intentionally omits gradients of the routing function. Consequently, our results do not characterize router sensitivity, end-to-end robustness, or routing instability. Likewise, weighted PCA captures linear variance structure under fixed routing and input distributions, but does not measure intrinsic dimensionality or nonlinear representational geometry.

**Expected failure modes.** Based on this design, we expect the observed geometric signatures to weaken in several regimes: (i) when routing is nearly uniform and expert specialization is weak; (ii) when strong architectural coupling or shared normalization dominates expert-local transformations; (iii) when experts are explicitly trained for predefined tasks, causing specialization to reflect task structure rather than routing-induced partitioning; and (iv) in deep or multi-layer MoE stacks, where routing effects may compound or average out across layers.

Despite these limitations, the controlled setting studied here provides a clear mechanistic baseline for understanding how expert routing reshapes local function geometry and representation variance. We view this work as complementary to large-scale empirical studies, offering interpretable geometric insights that can inform future analyses of MoE behavior in attention-based architectures and pretrained language models. The **large-scale models (e.g., 70B-class)** predictions (Section 5) remain theoretical pending validation on proprietary-scale checkpoints

## 5 Discussion and Conclusion

This work examined Mixture-of-Experts (MoE) architectures through a geometric lens, focusing on how expert routing reshapes local function sensitivity and representation-space variance in a controlled setting. By combining Jacobian spectral analysis with weighted PCA, we aimed to isolate routing-induced effects that are difficult to observe directly in large-scale language models.

**Summary of findings:** Across routing regimes and random seeds, we observed three consistent geometric patterns. First, expert-local Jacobians exhibited smaller leading singular values and faster spectral decay than dense baselines, indicating reduced local sensitivity under the expert transformations. Second, cross-expert Jacobian alignment remained near zero under a cosine-similarity metric, suggesting limited overlap in dominant sensitivity directions across experts. Third, representation variance was redistributed under routing: expert-local representations showed higher effective rank under weighted PCA than dense representations, with routing sharpness modulating the degree of variance concentration. Taken together, these results support an interpretation of MoEs as implementing a soft partitioning of function space, where routing decomposes sensitivity across experts while controlling the internal variance structure of expert-local representations. As emphasized in Section 4, these conclusions pertain to local geometric properties under specific probes and do not imply global guarantees about expressivity, robustness, or performance.

**Interpreting routing sharpness:** Our comparison of Top-k and fully-soft routing highlights routing sharpness as a key factor shaping expert-local geometry. Sharper routing produces experts that operate on narrower input distributions, leading to more concentrated variance and lower effective rank under PCA. Softer routing exposes experts to more heterogeneous inputs, resulting in broader variance structure and flatter PCA spectra. Importantly, the presence of low cross-expert Jacobian alignment across both regimes suggests that functional decomposition arises even when routing is not sparse. Routing sharpness therefore appears to regulate *how* experts organize internal structure, rather than *whether* they specialize at all, at least in the setting studied here.

**Relation to prior MoE perspectives:** Most prior work on MoEs emphasizes efficiency, scalability, and routing stability. From that perspective, expert specialization is often inferred indirectly through load balancing or task performance. Our results complement this view by providing direct geometric probes of expert-local behavior, offering a mechanistic perspective on how specialization manifests at the level of local sensitivity and representation variance. At the same time, recent work on expert merging and MoE compression suggests that many experts can be approximated or combined without significant loss. Our findings do not contradict this observation: low cross-expert alignment under a Jacobian metric does not preclude functional redundancy under other metrics or tasks. Rather, the geometric decomposition we observe may coexist with approximate compressibility, depending on how expert behavior is measured.

**Implications and hypotheses:** While our analysis is intentionally limited in scope, it suggests several hypotheses that can be tested in more realistic settings. If similar geometric patterns persist in transformer-based MoEs, one might expect routing to influence robustness, optimization dynamics, or representation diversity by modulating local sensitivity and variance concentration. Conversely, architectural components such as attention and normalization may attenuate or reshape these effects. Our results also suggest that routing design choices—such as the degree of sparsity or softness—can have systematic effects on internal representation geometry, beyond their role in compute allocation. Understanding these effects may help inform routing strategies that balance specialization and overlap in practice. While our primary analysis is restricted to synthetic data and a shallow transformer, the geometric signatures validated on WikiText (Section 3.5) suggest the following testable hypotheses for deep, large-scale MoEs: (1) **The Geometric Elbow:** Given our interpretation of MoE routing as a form of soft-clustering, we hypothesize that expert-scaling follows a **clustering-like 'elbow' phenomenon**, where variance concentration identifies a point of diminishing returns for representational capacity; (2) **Hallucination Suppression:** A potential reduction in hallucination artifacts, as the observed **reduction in Jacobian norms** implies flatter local mappings that are less sensitive to the infinitesimal perturbations that often seed error accumulation in long-context generation; and (3) **Ensemble Reliability:** Enhanced ensemble reliability in Top-K over fully-soft routing, driven by near-orthogonal expert Jacobians that maximize **functional diversity** within the expert pool. These results provide actionable design criteria for balancing routing sharpness and representation rank in frontier long-context transformers.

**Outlook:** This work establishes a controlled geometric baseline for analyzing MoE architectures. Future work can extend these probes to deeper and multi-layer MoEs, incorporate router gradients into end-to-end Jacobian analysis, and apply similar techniques to attention-based and pretrained language models. Our observation that representation-space rank inverts between synthetic and real data while function-space curvature reduction remains stable suggests a geometric duality that warrants deeper theoretical investigation: **MoEs may flatten function space while compressing representation space, a trade-off that could inform routing design beyond simple load balancing**. Another promising direction is to connect geometric measures more directly to downstream behavior, such as robustness to perturbations or generalization across domains.

By clarifying the scope and limitations of our conclusions and by grounding interpretation in explicit geometric probes, we hope this work provides a useful reference point for future studies of expert routing and specialization.

## Appendix A  Notation, Formal Definitions, and Methodological Details

This appendix provides the full mathematical definitions and derivations underlying the notation introduced in Section 2.1. We include the formal routing definitions, weighted PCA construction, Jacobian estimation procedure, and notation tables used throughout the paper.

### A.1  MoE Layer and Routing Functions

An MoE layer consists of a set of experts

$$F = \{f_e: \mathfrak{R}^{d_{model}} \to \mathfrak{R}^{d_{model}}\}_{e=1}^{E}$$

And a routing function

$$g: \mathfrak{R}^{d_{model}} \to \Delta^{E-1}$$

Where $\Delta^{E-1}$ is the probability simplex.

For an input hidden state $x \in \mathfrak{R}^{d_{model}}$, the MoE output is

$$f_{MoE}(x) = \sum_{e=1}^{E} g_e(x) f_e(x)$$

**Top-k routing.**

The router produces a sparse vector $g_e(x)$, let $TopK(x) \subseteq \{1,...,E\}$ denote the selected experts, then

$$g_e(x) = \left\{ \frac{exp(s_e(x))}{\Sigma_{j \in TopK(x)} exp(s_j(x))},\ e \in TopK(x),\ o,\ otherwise \right\}$$

Where $s_e(x)$ is the router logits.

**Fully-soft routing.**

All experts receive non-zero weight:

$$g_e(x) = \frac{exp(s_e(x))}{\Sigma_{j=1}^{E} exp(s_j(x))}$$

### A.2  Expert-Local Jacobians

For each expert $e$, the **expert-local Jacobian** is

$$J_e(x) = \frac{\partial f_e(x)}{\partial x} \in \mathfrak{R}^{d_{model} \times d_{model}}$$

The Jacobian of the full MoE layer is:

$$J_{MoE}(x) = \frac{\partial}{\partial x}\left(\sum_{e=1}^{E} g_e(x) f_e(x)\right) = \sum_{e=1}^{E} g_e(x) J_e(x) + \sum_{e=1}^{E} f_e(x) \frac{\partial g(x)}{\partial x}$$

In this work, we focus on **expert-local geometry**, analyzing $J_e(x)$ rather than the full MoE Jacobian $J_{MoE}(x)$. This isolates the functional structure learned by each expert independent of router gradients.

**Jacobian spectra.**

For each expert, we compute the singular values

$$\sigma_1(J_e(x)) \geq \sigma_2(J_e(x)) \geq ... \geq \sigma_r(J_e(x))$$

We report:

- The largest singular value $\sigma_1(J_e(x))$, which is also called the Jacobian norm, the spectral norm, Lipschitz constant and worst-case sensitivity.
- normalized cumulative energy

The average Jacobian for expert $e$ is

$$\bar{J}_e = \frac{1}{N_e} \sum_{i=1}^{N} g_e(x_i) J_e(x_i)$$

Where $N_e = \sum_{i=1}^{N} g_e(x_i)$ is the effective sample count.

### A.3 Cross-Expert Jacobian Similarity

To quantify functional decomposition, we compute cosine similarity between average expert Jacobians of two experts $e_1$, $e_2$:

$$cos(e_1, e_2) = \frac{<vec(\bar{J}_{e_1}), vec(\bar{J}_{e_2})>}{\left\|vec(\bar{J}_{e_1})\right\|_2 \left\|vec(\bar{J}_{e_2})\right\|_2}$$

Near-zero similarity indicates that experts implement **distinct transformation subspaces**.

### A.4 Weighted PCA for Routed Hidden States

Let $\{h_i\}_{i=1}^{N}$ be hidden states entering the MoE layer. For expert $e$, define the expert-weighted samples:

$$h_{i,e} = \sqrt{g_e(h_i)} h_i$$

The weighted covariance matrix is:

$$C_e = \frac{1}{N_e} \sum_{i=1}^{N} g_e(h_i)(h_i - \mu_e)(h_i - \mu_e)^T$$

Where $\mu_e = \frac{1}{N_e} \sum_{i=1}^{N} g_e(h_i) h_i$

We compute eigenvalues

$\lambda_{1,e} \geq \lambda_{2,e} \geq \cdots \geq \lambda_{d,e}$

The Explained variance is

$$e_i = \frac{\lambda_i}{\Sigma_j \lambda_i}$$

Cumulative variance is:

$$E_k = \sum_{i=1}^{k} e_i = \frac{\sum_{i=1}^{k} \lambda_i}{\Sigma_j \lambda_j}$$

We use the number of components required to reach 90% cumulative variance (*k*@0.9) as an effective rank metric. This construction ensures that each expert's PCA reflects the distribution of inputs it actually receives under the router.

### A.5 Implementation Details

2. The largest singular value $\sigma_1$ of Jacobians is extracted via exact SVD. For deeper models where J becomes prohibitively large, $\sigma_1$ can be estimated via power iteration.
3. PCA computed on centered, weighted hidden states.
4. Top-k uses k=2 unless otherwise specified.
5. Fully-soft uses full softmax over experts.

## Appendix B   Experimental Setup

This appendix describes the model configurations, datasets, routing settings, and probing procedures used in our spectral geometry analysis of Mixture-of-Experts (MoE) models.

### B.1   Models

We evaluate three model classes:

- **Dense Transformer (baseline).** A standard feedforward block with identical hidden dimension, activation, and parameter count to the MoE experts, but without routing.
- **MoE-Top-k.** A sparsely-activated MoE layer using Top-k routing with k=2 unless otherwise specified.
- **MoE-Fully-Soft.** A fully-soft MoE layer using a softmax router over all experts.

All models share the same hidden dimension, number of heads, feedforward width, and activation functions. The only difference across conditions is the routing mechanism.

**Expert configuration**

1. Number of experts: E=8
2. Expert architecture: 2-layer MLP with GELU
3. **Model dimension:** $d_{model}$=64
4. Expert hidden dimension (same as dense FFN): $d_{hidden}$=128
5. **Batch size:** 512 tokens
6. Router: single linear projection followed by softmax or Top-k selection

This configuration provides a balanced regime where (i) routing behavior is non-trivial, (ii) Jacobian computation remains tractable, and (iii) PCA spectra are stable across seeds. The dimensions are chosen to match the per-expert capacity of small transformer feedforward blocks while enabling full spectral analysis without approximation.

### B.2 Synthetic Data Generation

To isolate the geometric effects of routing from a dataset-specific structure, all experiments use fixed synthetic inputs rather than natural language corpora. This controlled setting ensures that differences in PCA spectra, Jacobian curvature, and expert specialization arise solely from the routing mechanism and not from semantic or distributional artifacts.

We generate a batch of hidden states

$$x \sim \aleph(0, I_{d_{model}})$$

using a fixed random seed for reproducibility. Targets are sampled independently from the same distribution:

This synthetic-data regime provides a stable foundation for computing full Jacobian spectra and weighted PCA without the confounding influence of tokenization, sequence structure, or linguistic variability. Because the goal of this work is to analyze **spectral geometry** rather than downstream performance, no tokenization or text preprocessing is required.

### B.3 Routing Settings

**Top-k routing**

For the Top-k MoE configuration, the router produces a sparse distribution over experts:

A. **Top-k:** k=2
B. **Routing weights:** obtained by applying a softmax to router logits, then selecting the top-k entries
C. **Temperature:** 1.0
D. **Noisy gating:** disabled
E. **Normalization:** selected expert weights are renormalized to sum to 1

This setting enforces sharp expert selection and produces sparse expert activation patterns.

**Fully-soft routing**

For the fully-soft MoE configuration, the router assigns non-zero weight to all experts:

- **Routing weights:** full softmax over all E experts
- **Temperature:** 1.0
- **No sparsity constraints:** every expert contributes to the output
- **No renormalization:** softmax already produces a valid probability distribution

This setting yields diffuse expert activation and higher-entropy routing behavior.

**Batching and weighting**

Routing weights $g_e(x)$ are computed per token (per sample in the synthetic batch). These weights are used consistently across all probes:

- Weighted PCA: routing weights serve as sample weights for expert-specific covariance estimation
- Jacobian averaging: expert-local Jacobians are averaged using the same per-token routing weights

This ensures that each expert's spectral geometry reflects the distribution of inputs it actually receives under the router.

### B.4 Jacobian Computation

To analyze expert-local function geometry, we compute the Jacobian of each expert with respect to the input hidden state. Because our experiments use a controlled MLP-MoE architecture with synthetic inputs, we can compute full Jacobian matrices without approximation.

For each sample $x$ and expert $e$, we compute

$$J_e(x) = \frac{\partial f_e(x)}{\partial x}$$

using automatic differentiation. Expert-local Jacobians are then aggregated using routing weights:

$$\bar{J}_e = \frac{1}{N_e} \sum_{i=1}^{N} g_e(x_i) J_e(x_i), \quad N_e = \sum_{i=1}^{N} g_e(x_i)$$

We analyze:

- singular-value spectra of $\bar{J}_e$
- largest singular values (curvature)
- spectral decay (distribution of curvature across directions)
- cross-expert cosine similarity (functional decomposition)

All Jacobians are computed on a fixed synthetic batch to ensure reproducibility and eliminate dataset-induced variability.

### B.5 Weighted PCA Procedure

To probe expert-local representation geometry, we perform weighted PCA on the hidden states entering the MoE layer. For each expert $e$, routing weights $g_e(x)$ determine the contribution of each sample.

Given hidden states $\{h_i\}_{i=1}^{N}$, we compute:

- Weighted mean

$$\mu_e = \frac{1}{N_e} \sum_{i=1}^{N} g_e(h_i) h_i$$

1. Weighted covariance

$$C_e = \frac{1}{N_e} \sum_{i=1}^{N} g_e(h_i)(h_i - \mu_e)(h_i - \mu_e)^T$$

- Eigenvalue decomposition

$$C_e = V_e \Lambda_e V_e^T$$

where $\Lambda_e = diag(\lambda_{1,e}, \lambda_{2,e}, \cdots \lambda_{d,e})$

We report:

12. eigenvalue spectra
13. cumulative variance curves
14. cum-var@10
15. effective rank

Weighted PCA ensures that each expert's representation geometry reflects the distribution of inputs it actually receives under the router.

**B.6 Training Details**

Both the dense baseline and the MoE models are trained using mean-squared error (MSE) on fixed synthetic inputs. We use the Adam optimizer with a learning rate of $1\times10^{-3}$. Each model is trained independently using the same batch of synthetic data to ensure that differences in spectral geometry arise solely from the routing mechanism.

The training loop consists of forward passes through the dense model and the MoE model, followed by backpropagation and parameter updates.

Because the architecture is lightweight and the dataset is synthetic, full Jacobian matrices can be computed exactly after training. All spectral analyses (Jacobian spectra, PCA spectra, and cross-expert similarity) are performed on the trained models using the same fixed batch of synthetic inputs.

**B.7 Hardware**

All experiments are executed on a CPU-only environment. The controlled MLP-MoE architecture and synthetic-data setup make full Jacobian computation tractable without GPU acceleration. Running on CPU ensures deterministic behavior across seeds and platforms, and avoids variability introduced by GPU kernel implementations.

**Appendix C  Extended Probes on a 3-layer Transformer with WikiText**

This appendix summarizes the **extended probes on a 3-layer Transformer with WikiText.**

## C.1 Jacobian norm comparison between dense, fully soft routing and top-2 routing models

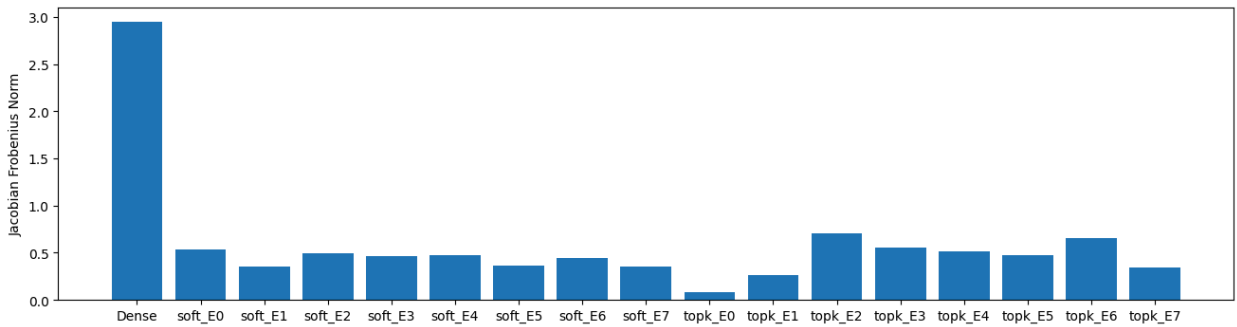

**Figure C.1:** Jacobian norms for dense, fully soft routing, and top-2 routing models. Routing-based models exhibit reduced global Jacobian magnitude relative to the dense baseline.

## C.2 Cross-expert Jacobian cosine similarities for fully soft routing and top-2 routing models

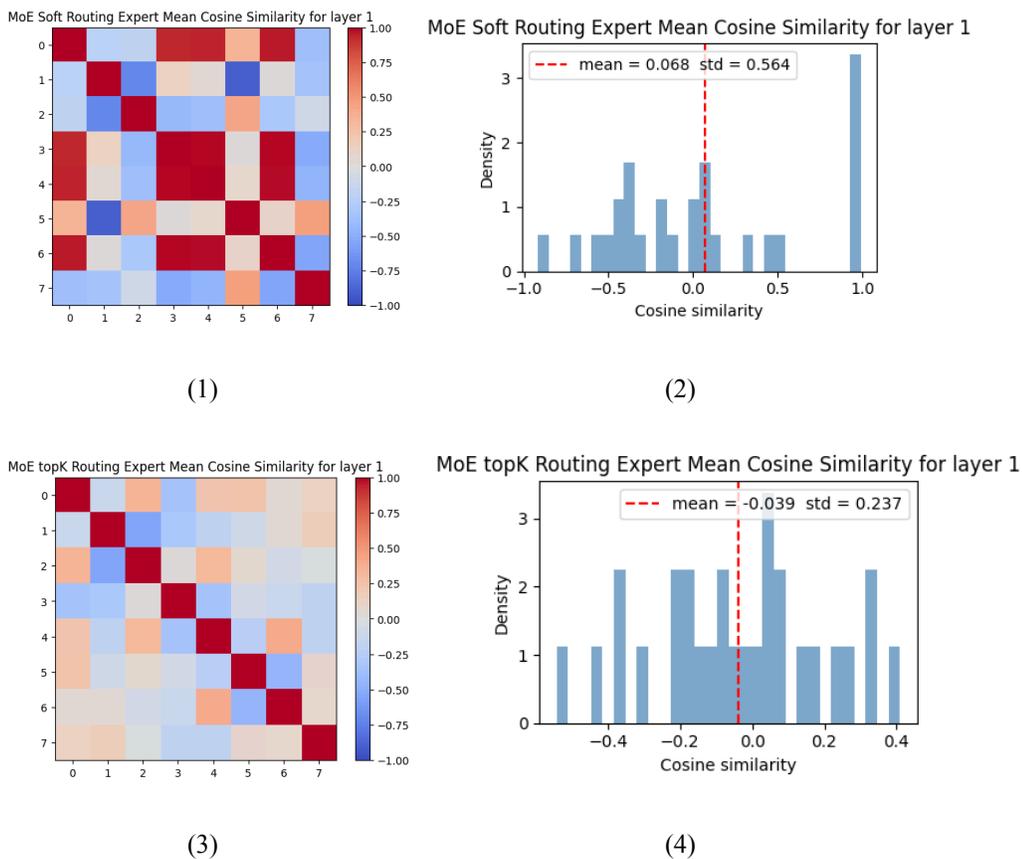

(1)  (2)

(3)  (4)

**Figure C.2:** Heatmaps and distribution plots of cross-expert Jacobian cosine similarities. Panels **(1)**–**(2)** correspond to the fully soft routing model, while panels **(3)**–**(4)** correspond to the top-2 routing model. Fully soft routing exhibits higher cross-expert similarity, whereas top-2 routing yields more orthogonal expert Jacobians, indicating stronger geometric specialization.

## C.3 PCA Spectrum and Cumulative Explained-variance Curves

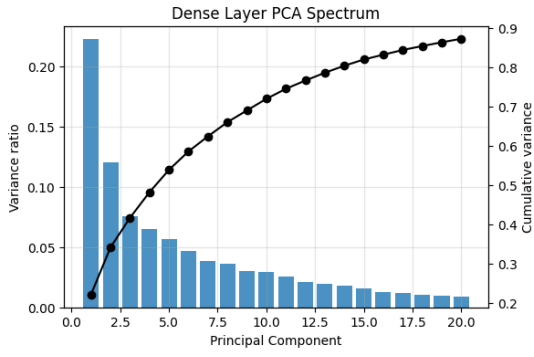

(1)

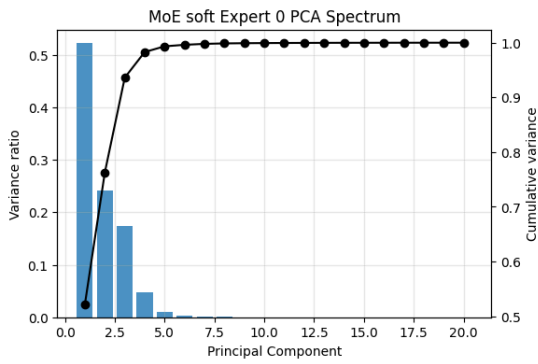

(2)

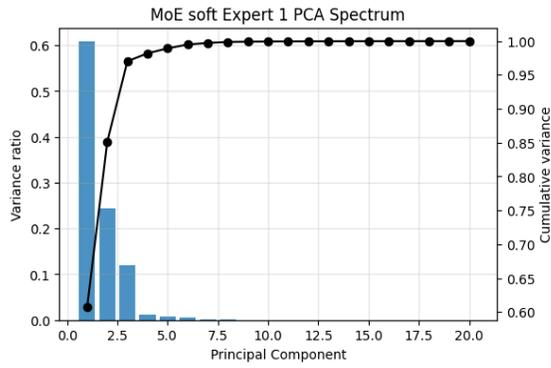

(3)

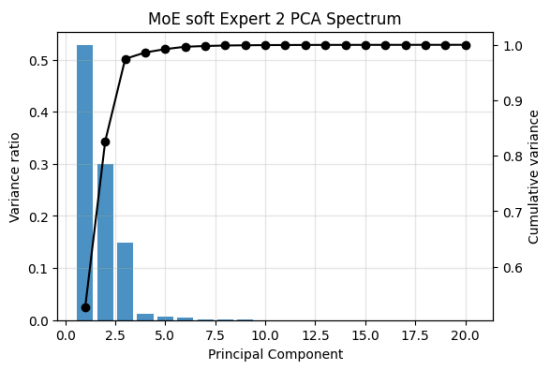

(4)

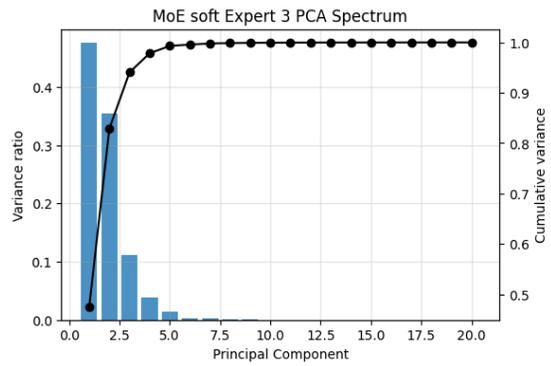

(5)

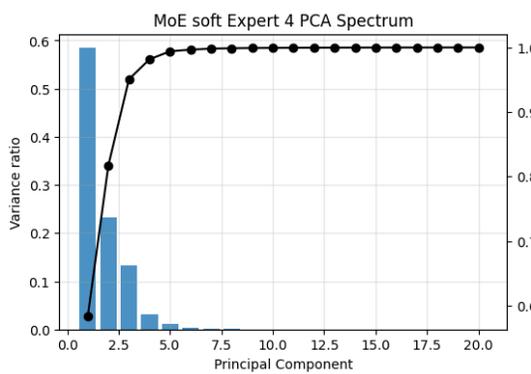

(6)

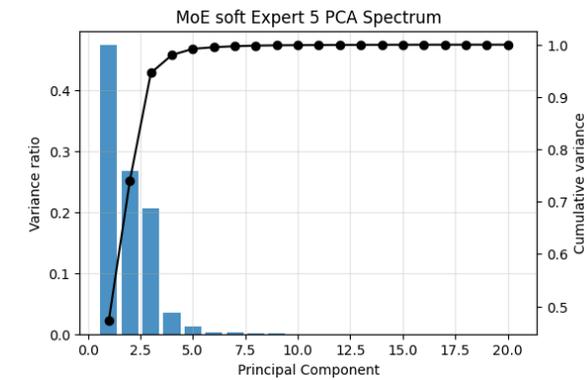

(7)

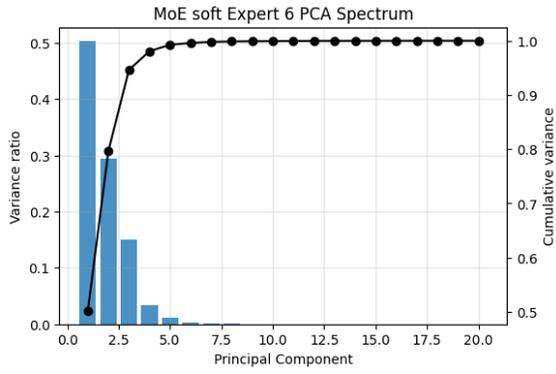
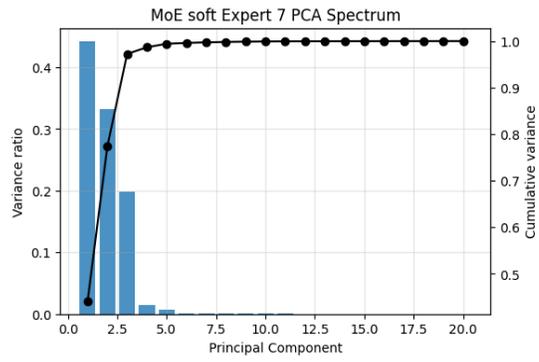

(8)  (9)

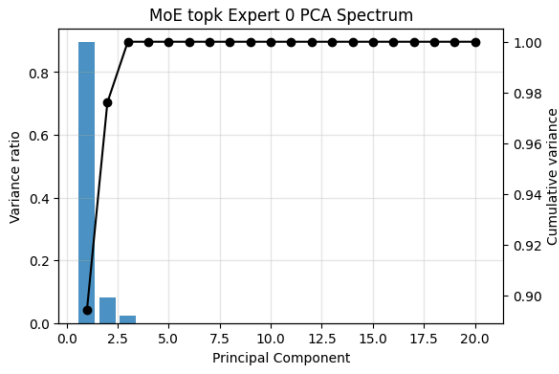
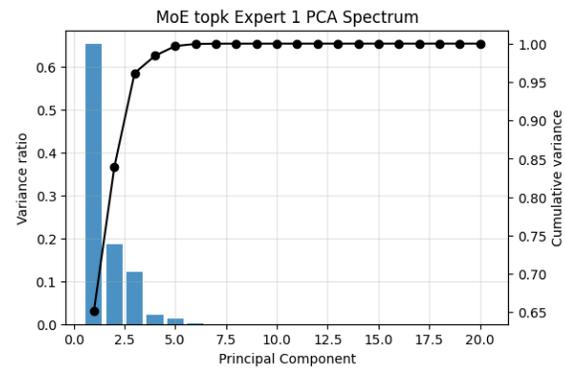

(10)  (11)

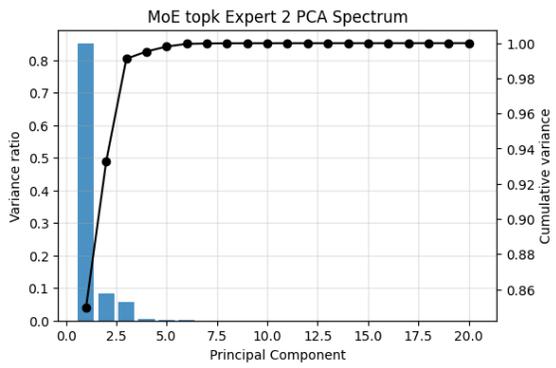
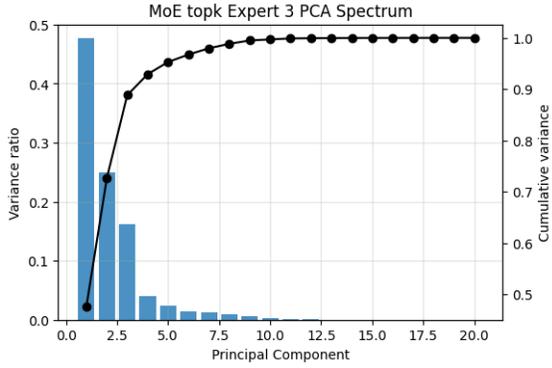

(12)  (13)

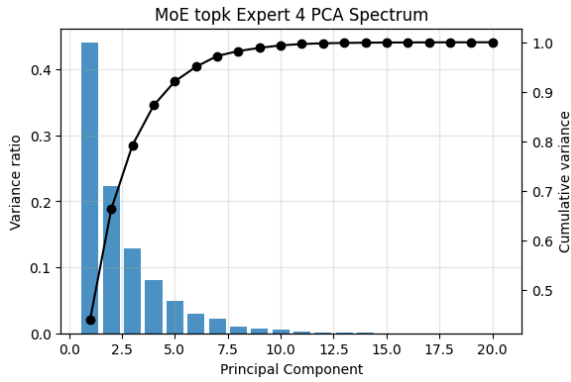

(14)

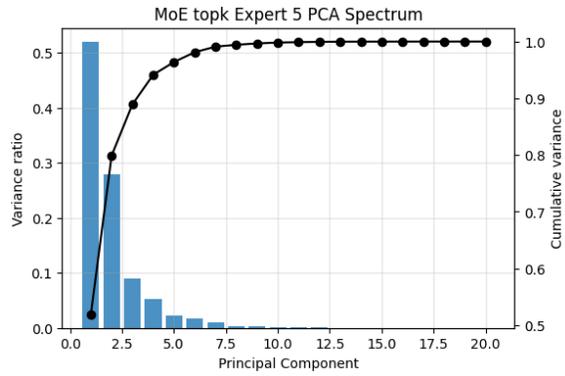

(15)

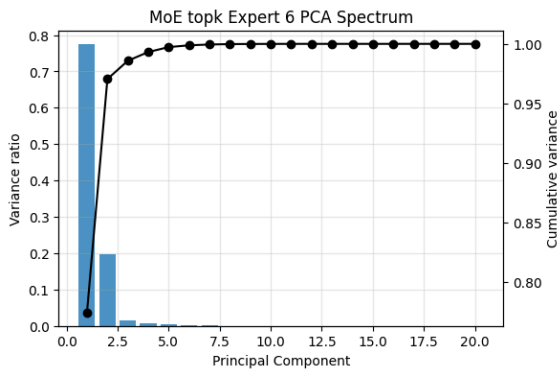

(16)

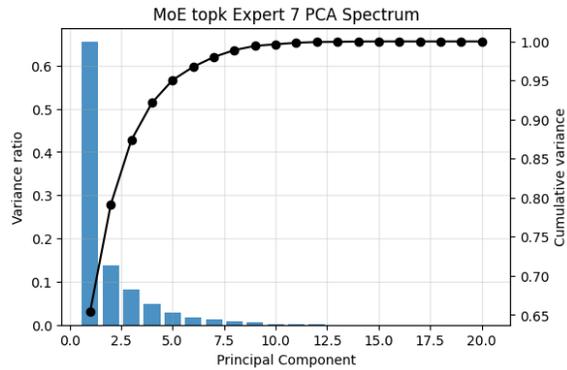

(17)

**Figure C.3:** Weighted PCA spectra and cumulative explained-variance curves. Panel **(1)** corresponds to the dense model; panels **(2)–(9)** show the eight experts of the fully soft routing model; and panels **(10)–(17)** show the eight experts of the top-2 routing model. The dense FFN requires approximately 20 principal components to explain 90 % of the variance, while each MoE expert reaches the same threshold with roughly 3 components, indicating substantially lower intrinsic dimensionality at the expert level.